\documentclass[letterpaper]{article} 
\usepackage{aaai25}  
\usepackage{times}  
\usepackage{helvet}  
\usepackage{courier}  
\usepackage[hyphens]{url}  
\usepackage{graphicx} 
\usepackage{natbib}  
\usepackage{caption} 
\usepackage{algorithm}
\usepackage{algorithmic}
\usepackage{amsmath}
\usepackage{amsfonts,amssymb}
\usepackage{booktabs}
\usepackage{multirow}

\usepackage{newfloat}
\usepackage{listings}
\DeclareCaptionStyle{ruled}{labelfont=normalfont,labelsep=colon,strut=off} 
\floatstyle{ruled}
\newfloat{listing}{tb}{lst}{}
\floatname{listing}{Listing}
\title{GRAIN: Multi-Granular and Implicit Information Aggregation \\ Graph Neural Network for Heterophilous Graphs}

\author{
    Songwei Zhao\textsuperscript{\rm 1,\rm 2},
    Yuan Jiang\textsuperscript{\rm 3}\thanks{Corresponding author.},
    Zijing Zhang\textsuperscript{\rm 1,\rm 2},
    Yang Yu\textsuperscript{\rm 1,\rm 2},
    Hechang Chen\textsuperscript{\rm 1,\rm 2}\footnotemark[1]
}
\affiliations{
    \textsuperscript{\rm 1}School of Artificial Intelligence, Jilin University, Changchun\\
    \textsuperscript{\rm 2}Engineering Research Center of Knowledge-Driven Human-Machine Intelligence, Jilin University, Changchun \\
    \textsuperscript{\rm 3}School of Computer Science and Engineering, Nanyang Technological University, Singapore \\

    \{zhaosw22, zijing23, yyu23\}@mails.jlu.edu.cn, yuan005@ntu.edu.sg, chenhc@jlu.edu.cn
}

\usepackage{bibentry}

\begin{document}

\maketitle

\begin{abstract}

Graph neural networks (GNNs) have shown significant success in learning graph representations. However, recent studies reveal that GNNs often fail to outperform simple MLPs on heterophilous graph tasks, where connected nodes may differ in features or labels, challenging the homophily assumption. Existing methods addressing this issue often overlook the importance of information granularity and rarely consider implicit relationships between distant nodes. To overcome these limitations, we propose the Granular and Implicit Graph Network (GRAIN), a novel GNN model specifically designed for heterophilous graphs. GRAIN enhances node embeddings by aggregating multi-view information at various granularity levels and incorporating implicit data from distant, non-neighboring nodes. This approach effectively integrates local and global information, resulting in smoother, more accurate node representations. We also introduce an adaptive graph information aggregator that efficiently combines multi-granularity and implicit data, significantly improving node representation quality, as shown by experiments on 13 datasets covering varying homophily and heterophily. GRAIN consistently outperforms 12 state-of-the-art models, excelling on both homophilous and heterophilous graphs.

\end{abstract}

\section{Introduction}\label{sec:introduction}

Graph Neural Networks (GNNs) \cite{welling2016semi} are a specialized class of deep neural networks designed to process and analyze graph-structured data. GNNs capitalize on the inherent properties of graphs, where entities are represented as nodes and their relationships as edges, to effectively capture complex interdependencies between entities. By employing iterative message-passing and aggregation mechanisms, GNNs iteratively update each node's representation by combining its features with those of its neighbors. This process enables GNNs to learn sophisticated and informative embeddings that are highly effective for a variety of graph-based machine learning tasks, such as node classification \cite{he2024improving}, link prediction \cite{lu2023schema}, and graph classification \cite{zhao2024twist}, often surpassing the performance of traditional neural networks.
GNNs have also demonstrated remarkable success across a broad spectrum of real-world applications, including social network analysis \cite{zhang2022improving}, recommendation systems \cite{agrawal2024no}, and drug discovery \cite{liu2023interpretable}. However, the primary reason GNNs excel in many tasks—their reliance on the homophily assumption—also presents a significant limitation. The homophily assumption presupposes that connected nodes tend to share similar attributes or labels, providing additional context for information aggregation. While this assumption works well in homophilous graphs, it fails to capture the complexities of heterophilous graphs, where connected nodes may have dissimilar features or labels.

As a result, increasing research shows that GNNs do not always outperform traditional deep neural networks in graph tasks, particularly in heterophilous settings. In some cases, even simple multi-layer perceptrons (MLPs) can surpass GNNs \cite{liu2021non,chien2020adaptive}. This discrepancy is largely due to the heterophily problem, where the homophily assumption breaks down, and GNNs struggle to effectively aggregate information from connected nodes with differing attributes. Recently, scholars have increasingly focused on heterophily and proposed models to address this issue. For instance, \citeauthor{chanpuriya2022simplified} \shortcite{chanpuriya2022simplified} proposes the Adaptive Simple Graph Convolution, which selects different filters for each feature, demonstrating its adaptability to both homophilous and heterophilous graph structures in experiments. 
GBK-GNN \cite{du2022gbk} captures both homophilous and heterophilous information through bi-kernel feature transformation and introduces a selection gate to choose the appropriate core for a given node.
\citeauthor{xiao2023spatial} \shortcite{xiao2023spatial} addresses the spatial heterophily in urban graphs by designing a rotation-scaling spatial aggregation module and a heterophily-sensitive spatial interaction module. LRGNN \cite{liang2024predicting} tackles label relation prediction in heterophilous graphs by solving an approximation problem of the global label relation matrix for signed graphs, making the proposed model applicable in both homophilous and heterophilous settings. 

Despite achieving good results in addressing heterophily, existing works generally underperform on homophily \cite{chien2020adaptive}. Moreover, they often overlook the importance of aggregation at various granularity levels in node representations, where coarse-grained information captures the overall position or influence of nodes (e.g., users) within the network, while fine-grained information focuses on their direct interactions with closely connected nodes.
Additionally, these methods rarely consider implicit relationships between distant nodes that are not neighbors but may share common interests or characteristics. Such implicit information is crucial for generating smooth embeddings by capturing underlying connections beyond neighbors.

Incorporating the aforementioned multi-view information into the aggregation process presents significant challenges. First, real-world graph data is inherently complex and diverse, with varying attributes and types, making it difficult to accurately determine the appropriate range of coarse and fine-grained information for each node. Second, even if the correct granularity levels can be identified, the task of effectively aggregating and training nodes within these ranges remains daunting. Different nodes necessitate tailored granularity ranges, which implies the need for varying network depths during training, thereby substantially increasing model complexity. Lastly, integrating implicit relationships between distant nodes into the aggregation process is a crucial challenge that remains unresolved.

To tackle the challenges of representing heterophilous graphs, we introduce the Granular and Implicit Graph Network (GRAIN), which achieves smoother and more comprehensive representations through multi-granularity information aggregation. First, we model graph representation learning as a Markov Decision Process (MDP), enabling systematic optimization of granularity selection based on long-term rewards, and efficiently incorporate Reinforcement Learning (RL) methods to solve this problem. Our RL framework is specifically designed to output continuous values, enabling flexible adjustment of granularity levels for different nodes. This customization allows for tailored information aggregation that seamlessly integrates both local and global perspectives, thereby enhancing the depth and richness of the node embeddings. Additionally, we introduce a novel node information aggregator that synergizes multi-granularity with implicit relationships between distant nodes, resulting in more refined and accurate node representations.

To validate the effectiveness of the proposed model, we conducted extensive experiments on 13 datasets with varying homophily rates, comparing GRAIN with 12 SOTA models. The results show that GRAIN not only performs excellently on homophilous datasets but also demonstrates superior performance on heterophilous datasets. The key contributions of this paper are as follows:
\begin{itemize}
    \item [-]  We propose a novel GNN model that integrates multi-granularity information and implicit relationships, achieving smoother and more accurate node embeddings across a variety of graph structures.
    
    \item [-] We devise two core components: Intelligent Granularity Perceiver, which systematically explores various granularities and implicit information while optimizing long-term rewards, and Multi-view Aggregator, which integrates this information into node representations for a cohesive synthesis of diverse data perspectives.

    \item [-] Extensive experiments on 13 benchmark datasets show that GRAIN excels in both homophilous and heterophilous settings, significantly outperforming 12 state-of-the-art methods and demonstrating strong generalization and broad applicability.
\end{itemize}

\section{Preliminaries}

\textbf{Formal problem definition.}
Suppose that $\mathcal{G} = \left\{ \mathcal{V},\mathcal{E} \right\}$ represents an undirected graph,
where $\mathcal{V}=\left \{v_1, v_2,\cdots, v_n \right\}$ is the set of nodes with $\left|\mathcal{V}\right|=n$, and $\mathcal{E} \subseteq \mathcal{V} \times \mathcal{V}$ denotes the set of undirected edges. Each edge $e_{ij} \in \mathcal{E}$ indicates a connection between node $v_i$ and node $v_j$. We use ${\bf A}_{ij}$ to denote the adjacency matrix of graph $\mathcal{G}$, where ${\bf A}_{ij} = 1$ if there is an edge between $v_i$ and $v_j$, and 0 otherwise. We denote $\mathcal{N}_k(v)$ as the $k$-th order neighbors of node $v$. Assume each node has $d$-dimensional features, then the feature vectors of all nodes are represented as $\bf X \in \mathbb{R}^{n \times d}$, and their labels $\bf Y$ are composed of one-hot vectors. The objective of graph representation learning is to map the input graph features into a low-dimensional vector space. According to the message-passing mechanism of GNNs, our goal is to learn node representations by considering the coarse and fine-grained information within the $k$-th order neighbors $\mathcal{N}_k(v)$ and the implicit information between nodes. The learned embeddings are thus more likely to be smooth and preserved in the low-dimensional vector space, making them ready for downstream tasks.

Based on the above definitions of the notations, we define the aggregation optimization problem in graph representation learning as follows. Given a graph network, we model the aggregation process of this network as a MDP. We aim to optimize this process to learn an information granularity selection policy $\pi$. This policy enables the network to dynamically learn and adjust the information granularity while also capturing the implicit information between nodes, thereby improving the overall performance of GNNs.

\noindent \textbf{Graph Neural Network.} Many popular GNN variants have been proposed to achieve smooth node representations effectively. One of the most prominent and widely used variants is the Graph Convolutional Network (GCN) \cite{kipf2016semi}.
GCNs aggregate information from a node's neighbors to update its representation. At each layer, a node's features are combined with those of its neighbors, using normalized connections to account for different node degrees. This aggregated information is then transformed through learnable weights and passed through an activation function like ReLU. The process iteratively refines the node's embedding, capturing both local features and structural information from the graph. This allows GCNs to effectively learn meaningful representations in graph-structured data.

\noindent \textbf{Reinforcement Learning for MDP.}
Reinforcement learning algorithms, such as Q-learning and policy gradient methods, commonly solve MDP problems by learning the optimal policy through environmental interactions. This study aims to consider coarse-grained information and implicit information between nodes, such as long-term dependencies and potential connections. Therefore, we utilize an Actor-Critic type RL algorithm to learn and adjust attention to different granularities of information. Additionally, we use the Twin Delayed Deep Deterministic Policy Gradient (TD3) algorithm \cite{fujimoto2018addressing}, designed for continuous action spaces, to learn the implicit information between nodes, thereby enhancing the expressive capacity of embeddings, detailed in the Appendix A.2.
In order to avoid overestimating the true value, the method takes the minimum between the two target networks, which gives the target update of the algorithm:
\begin{equation}
        target = reward + \gamma \cdot \min_{i \in \{1, 2\}} Q_{\theta^\prime_i} \left( s^\prime, a^\prime \right),
        \label{target}
\end{equation}
where $\gamma$ denotes the discount factor, $\theta^\prime$ denotes the parameters of target the critic network, $Q$ refers to the Q-value estimated by the target critic network, $s^\prime$ and $a^\prime$ are the next state and action, respectively.

While we employ TD3 to solve the MDP in graph representation learning in this study, we also compare value-based and policy-based deep RL methods in Appendix C.3. The experimental results highlight the advantages of the Actor-Critic framework in addressing our problem, particularly in its ability to explore implicit information between graph nodes. These findings further validate the flexibility of our proposed framework.

\section{Methodology} \label{method}

\begin{figure*}[t]
\centering 
    \includegraphics[width=14.7cm]{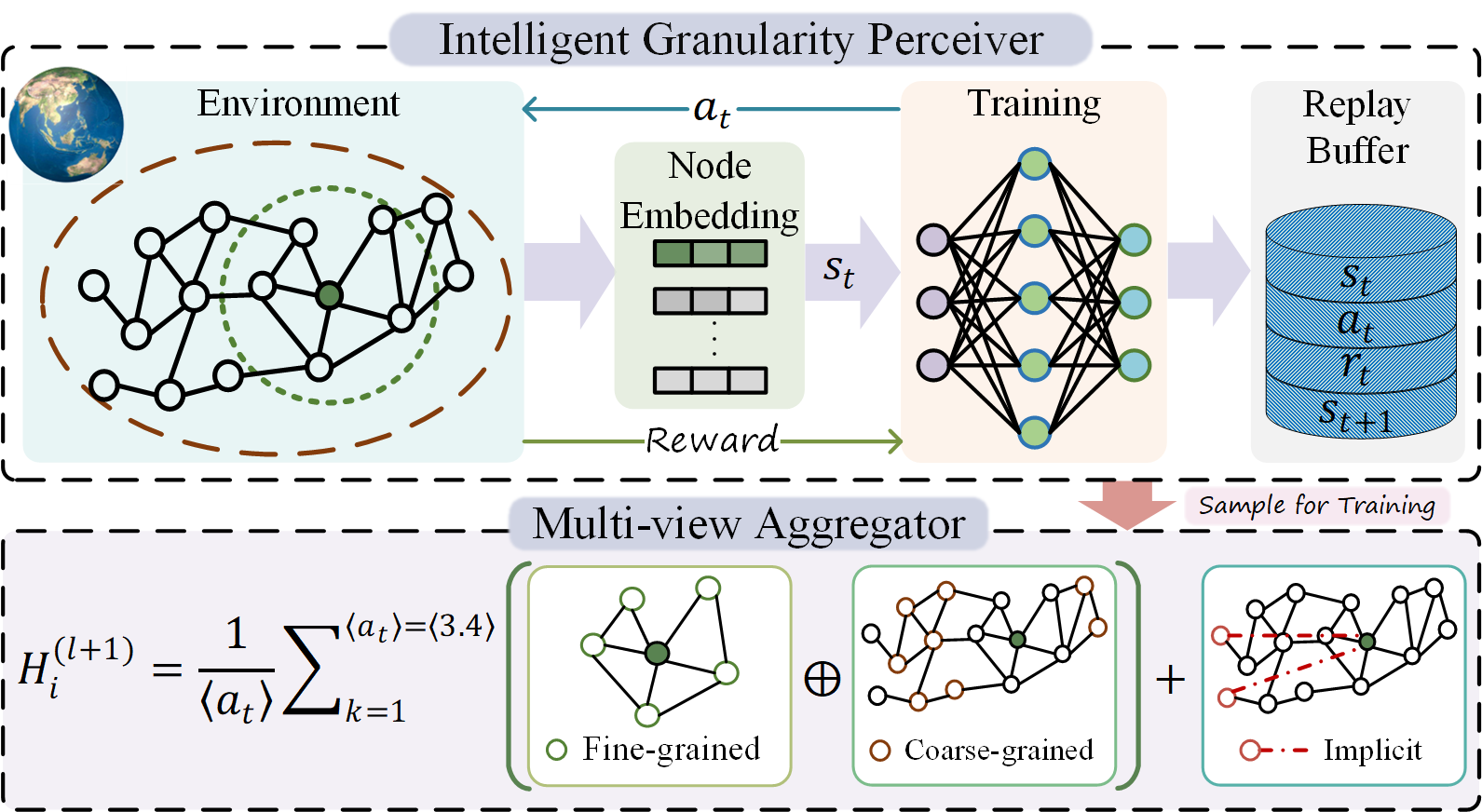}
    \caption{ An illustration of our proposed framework. The key idea of the model is to explore different levels of granularity information and implicit information of the target nodes in the graph through the intelligent granularity perceiver. We aim to smooth the representation by aggregating the different information into embedding the nodes through the multi-view aggregator.}
    \label{framework}
\end{figure*}

Figure \ref{framework} illustrates the overall framework of our proposed GRAIN. Our framework consists of two main components: the Intelligent Granularity Perceiver Module and the Multi-view Aggregator Module. The perceiver automatically learns and adjusts attention to different granularities of information from a global perspective. The aggregator utilizes the learned policies to integrate different granularities of information, enhancing the embedding of nodes. The algorithm's pseudo-code is presented in the Appendix B.


\subsection{Intelligent Granularity Perceiver}
\label{adaptive}
We discuss how to formulate the process of learning the optimal information granularity selection policy as a MDP. The critical components of an MDP include states, actions, rewards, and the transfer probability from the current state to the next state. Therefore, we need to define these components in the context of the graph network: 1) \emph{State} ($\mathcal{S}$):  In the context of our graph network, we define the state $s_{t} \in \mathcal{S}$ as the current representation of the node at the $t$-th iteration. 
2) \emph{Action} ($\mathcal{A}$): The action $a_{t} \in \mathcal{A}$ of the $t$-th iteration is defined as the number of hops for the current node to aggregate information. This action determines the scope of the neighborhood information that will be considered when updating the node's representation. 3) \emph{Reward function} ($\mathcal{R}$): The reward $r_{t}$ at the $t$-th iteration is defined as the average performance of the predictive task over the past few steps.

Based on the above definitions, we model the input graph network as an interactive environment. The process of information granularity selection involves the following three steps: 1) \emph{Select the Initial Node}: From the input graph data, select a starting node and use its features as the current state $s_{t}$. 2) \emph{Generate Action}: Utilize the RL policy to generate an action $a_{t}$ to guide the node in aggregating information from its neighbors, which determines the number of hops for the current node's aggregation.
3) \emph{Sample}: Sample a neighbor within the $a_{t}$ hops as the next time step's node, and use its input features as the next state $s_{t+1}$ in the MDP. Through the described process, we can effectively represent the node embedding learning process in the graph as a MDP.

It is challenging to automatically learn and adjust the attention to different information granularities for each node as graph data becomes more complex.
Implicit relationships between nodes are crucial for node representation quality, but integrating this information during aggregation remains a significant challenge.
RL algorithms \cite{mnih2013playing,mnih2015human}, with their decision-making processes based on environmental feedback, are well-suited for complex and dynamic environments where traditional optimization algorithms struggle to achieve optimal solutions. Therefore, we propose leveraging RL to optimize this process. To account for the implicit information between nodes and facilitate the model's better understanding of the relationships and semantic similarities among nodes, we employ an Actor-Critic RL algorithm to adapt attention to different levels of granularity in the information.
Additionally, we consider using the TD3 algorithm to capture implicit information between nodes, thereby enhancing node embedding.

The pivotal factor guiding the TD3 algorithm to learn the proposed MDP in the graph is the reward function, which we define as follows:
\begin{equation}
        \mathcal{R}\left(s_t,a_t\right)=\frac{\varphi \cdot \sum_{l=t-\vartheta}^{t} \left[ \mathcal{F} \left(s_t,a_t\right) -\mathcal{F} \left(s_l,a_l\right)\right]} {\vartheta+1},
         \label{reward}
\end{equation}
where $\varphi$ is a hyperparameter that plays a role in determining the strength of the reward signal, $\vartheta$ specifies the number of historical steps considered in the optimization process,
and $\mathcal{F}$ denotes the evaluation metric for the graph node classification task. The purpose of calculating the reward function over a certain range of historical windows is to encourage the agent to jointly consider the performance of the most recent $\vartheta$ time steps, which can lead to better policy exploration.
During training, our framework learns using nodes from the train set and utilizes the node classification accuracy on the validation set as the evaluation metric.

Based on reward function \eqref{reward}, the target value is computed using Equation \eqref{target}, and the mean square error loss is calculated against the current Q value. 
Following this, we update the actor network to ensure more stable training.

\subsection{Multi-View Aggregator}

In this subsection, we discuss how information granularity selection strategies can be used to enhance node representation learning through a novel aggregation function. The proposed model explicitly incorporates this strategy by mapping node embeddings to continuous hop counts, as shown in the aggregator part of Figure \ref{framework}. This aggregation function enables the target node to consider different granularities and implicit information, forming a specialized GNN architecture. Unlike previous aggregators, our approach enables capturing more relevant information, offering excellent compatibility and flexibility, thereby enhancing graph representation learning.

To validate the effectiveness of the proposed framework, we integrate the information granularity selection strategy and novel aggregation function into a basic GCN, deliberately choosing this simpler architecture to isolate and clearly demonstrate the impact of our approach. By replacing the original fixed receptive field and aggregation function, our framework adaptively customizes the number of aggregation hops for each node at time step $t$, enabling nodes to gather information at varying granularities from a global perspective and capture implicit information from latent connections and semantic similarities. This adaptive approach enhances each node's ability to learn relevant features in its unique context. The proposed aggregation function is constructed as follows:
\[ h_v^1 = \sigma \left( \sum\nolimits_{u_1 \in \mathcal{N}_1 \left( v \right)} { {\hat{a}}_{u_1v} {\bf X}_u } \right), \]   
\[ \vdots \]
\[ h_v^{k = \langle a_t \rangle} = \sigma \left( \sum\nolimits_{u_k \in \mathcal{N}_k \left( v \right)} {{\hat{a}}_{u_k u_{k-1}} h_v^{k-1} } \right), \]
\[ h_v = \frac{1} {\langle a_t \rangle} h_v^{\langle a_t \rangle} + \left( a_t - \left\lfloor a_t \right\rfloor \right) \hat{a}_{u_k u_{k-1}}h_v^{k-1} \]
\[+ \left( \left\lceil a_t \right\rceil - a_t \right) {\hat{a}}_{u_{k+1} u_k} h_v^k, \]
\[ output = {\rm {log \_ softmax}} \left( \sigma \left( {\rm Dropout} \left( h_v \right) \right) \right) \]
where $h_v^k$ is the feature vector of node $v$ at the $k$-th layer,
${\bf X}_u$ denotes the direct neighbor representation vector of node $u$.
The aggregation hop count $k = \langle a_t \rangle$ is determined by the policy function $\pi$ at time step $t$. Since $a_t$ is continuous, we round to the nearest integer. 
The $\hat{a}_{u_1 v}$ denotes the normalized adjacency matrix for the $1$-hop neighbors of node $v$.
In the formula for obtaining $h_v$, the first term represents the integrated information in coarse and fine granularity, while the latter terms capture the implicit information of potential connections and semantic similarity with the target node.
The functions $\left\lfloor \cdot \right\rfloor$ and $\left\lceil \cdot \right\rceil$ represent the floor and ceiling functions, respectively.
This formulation allows each node to dynamically adjust the aggregated granularity information, and enhances the ability to learn implicit information.

\subsection{Efficient GNN Aggregation} 

\begin{table*}[!htbp]
    \centering
    \scalebox{0.93}{
    \begin{tabular}{c|c|ccc|ccccc|c} 
        \hline
        \toprule
         Type         &Method     & Pubmed          & Citeseer     & Cora                & Cornell            & Actor              & Wisconsin                       & Texas    & Chameleon    &  Avg. Drop ($\downarrow$)     \\
        \toprule
        \multirow{4}{*}{Tradition}    &MLP   & 81.02          & 73.44          & 76.72          & 83.02          & 37.02         & 83.78            & 81.82     & 42.64      & 11.32    \\ 
        &GCN  & 83.77          & 75.51          & 84.28    & 62.26        & 30.69       & 53.27              & 51.11     & 44.42      & 28.33         \\
        &GAT   & 85.80          & 77.91          & 84.92        & 72.31          & 31.84         & 65.01              & 69.81    & 44.63       & 17.01         \\
        &GraphSAGE   & 85.78          & 76.65          & 84.74       & 71.68        & 32.31         & 78.63           & 70.73   & 41.57     & 14.89     \\  \toprule
        \multirow{3}{*}{Homophily}  &MixHop & 86.16          & 78.18          & 85.89              & 76.71          & 35.39         & 77.51              & 73.17     & 45.71      & 11.47    \\  
        &APPNP & 85.98          & 78.79          & 85.04             & 75.61          & 32.13    & 73.75                &71.95    & 37.86      & 15.09      \\
        &GCNII & 86.92         & 77.64          & 85.56         & 76.92      & 32.66         & 71.25               & 74.93     & 42.23    & 13.62     \\   \toprule
        \multirow{5}{*}{Heterophily} &GPR-GNN & 86.76          & 76.69          & 85.57             & 82.93          & 35.66       & 80.17              & 79.27    & 51.42     & 7.66      \\ 
        &GGCN  & 86.88       & 76.72       & 85.25           & 88.89       & 37.81        & 82.64                & 83.33    & 55.16    & 4.37     \\
        &ACM-GNN & 86.91          & 79.27          & 86.72               & 86.11          & 37.61       & 88.25                 & 84.62    & 56.04     & 2.85       \\
        &FE-GNN & 86.68          & 79.81          & 85.41              & 87.78          & 36.92       & 88.54               & 87.69    & 53.61      & 2.69      \\
        &GNN-SATA & OOM          & 78.46          & 86.83           & 87.81      & \textbf{39.66}      & 87.62              & 86.59    & 56.12    & 2.36      \\       \toprule
        Ours   &GRAIN   & \textbf{87.04}          & \textbf{81.25}          & \textbf{88.52}          & \textbf{90.12}          & 38.89         & \textbf{89.01}               & \textbf{87.69}    & \textbf{56.43}       & --     \\ 
        \hline
        \toprule
    \end{tabular}
    }
    \caption{The average test classification accuracy (\%) across all methods on eight real-world datasets. Drop ($\downarrow$) denotes how much the performance of baselines drop relative to our method. The best results are in bold. OOM refers to ``out-of-memory''.}
    \label{compare2}
\end{table*}

In this section, we introduce key techniques to enhance the model's training efficiency. Specifically, we employ an adaptive information granularity selection strategy for each node, which, while effective, results in a time-intensive construction of GNN aggregation at each time step. This leads to significant training overhead, posing a challenge in implementing the proposed model. Given that the number of hidden units in each node's aggregation layer is $N$, the parameter count required for training at each step can be substantial, calculated as $\frac {N \cdot \left( 2 + \langle a_t \rangle \right) \left( \langle a_t \rangle + 1 \right)} {2}$. This large parameter count considerably impacts training efficiency.


To address this issue, we introduce a parameter sharing mechanism to reduce parameters. During aggregation, each node's layer shares a set of parameters, enabling reuse. The aggregation function is expressed in Equation \eqref{represent}, reducing the training parameters to $N \cdot \langle a_t \rangle$:
\begin{align}
        {\bf H}_i^{(l+1)}=\frac{1} {\langle a_t \rangle} \sum\nolimits_{k =1}^{\langle a_t \rangle} \left[ (1 - \alpha) \hat{{\bf A}}_i^k {\bf H}^{(l)} + \alpha {\bf H}_i  \right]  + \notag \\
        (a_t - \left\lfloor a_t \right\rfloor) \hat{{\bf A}}_i^{\langle a_t \rangle} {\bf H}^{(l)} + ( \left\lceil a_t \right\rceil - a_t ) \hat{{\bf A}}_i^{\langle a_t \rangle + 1} {\bf H}^{(l)},
         \label{represent}
\end{align}
which not only lowers storage requirements but also decreases the model's computational load, thereby enhancing training efficiency. ${\hat{\bf A }}$ equals ${\bf A+I}$, where ${\bf I} \in \mathbb{R}^{n\times n}$ is the identity matrix. We incorporate the representation of target node $i$ into the formula and use the parameter $\alpha$ to balance the aggregated and node representations. This approach ensures that the newly obtained representation reflects both macro- and micro-level relationships while preserving critical features during aggregation. The Equation \eqref{represent} corresponds to the three components depicted in the lower parts of Figure \ref{framework}. The first term in the summation represents the integration of coarse and fine granularities, while the sum of the latter two terms captures the implicit information between nodes. When dealing with a continuous action space, we take the floor of the action to determine the range of coarse-grained information. The implicit information denotes the data between nodes within the ceiling of the action range and the target node.

Beside parameter sharing, we introduce a buffering mechanism to further improve the model's efficiency. The buffering mechanism stores the states, actions, and rewards collected during exploration at each time step, avoiding redundant computations during training data collection. Once the buffer reaches the batch size, the data stored in the buffer is used to train the GNN, thus accelerating the inference process of the framework. By combining the buffering mechanism with parameter sharing, we achieve a dual enhancement of the model's efficiency, significantly improving the practicality of the proposed framework.



\section{Experiments}
\label{Experiments}
In this section, we conduct a comprehensive evaluation of the GRAIN algorithm by comparing it against 12 baseline methods across 13 diverse datasets. The analysis of the experimental results highlight the effectiveness and strong generalization ability of GRAIN.

\noindent \textbf{Datasets and Setup.}
The proposed GRAIN is evaluated on both homophilous and heterophilous datasets. For homophily data, we use three citation networks \cite{sen2008collective}, i.e., Cora, Citeseer, and Pubmed, two Amazon co-purchase networks \cite{shchur2018pitfalls}, specifically Computers and Photo, and one co-authorship network (Coauthor CS) \cite{shchur2018pitfalls}.
For heterophily data, we use seven network datasets \cite{pei2020geom}, including Cornell, Wisconsin, Texas, Film, Actor \cite{tang2009social}, Squirrel, and Chameleon.
For the GNN module, we set the batch size to 128 and use \emph{Relu} as the activation function. The complete details of datasets, experimental settings, and full results are provided in Appendix C.




\noindent \textbf{Comparative Algorithms.}
We compare GRAIN with 12 classical and advanced baseline methods to validate the effectiveness of the proposed approach. These methods include a two-layer MLP, classical GNN methods such as GCN \cite{welling2016semi}, GAT \cite{velivckovic2017graph}, and GraphSAGE \cite{hamilton2017inductive}; advanced methods designed for homophily: including MixHop \cite{abu2019mixhop} utilizes aggregation at different hop counts, APPNP \cite{gasteiger2018predict} introduces personalized information propagation, and GCNII \cite{chen2020simple} mitigates over-smoothing by using initial residuals and identity mapping. For handling heterophily, we compare with state-of-the-art methods such as GPR-GNN \cite{chien2020adaptive}, which adaptively learns the weights of Generalized PageRank to optimize features, GGCN \cite{yang2021graph}, which proposes structural and feature-based edge correction, ACM-GCN \cite{luan2021heterophily}, which adaptively exploits aggregation, diversification, and identity channels to address harmful heterophily, FE-GNN \cite{sun2023feature}, which constructs features using Chebyshev polynomials or monomials, and GNN-SATA \cite{yang2024graph}, which utilizes soft association to identify correlations between features and structure. 



\subsection {Comparative Results}
In this section, we first define the homophily rate to classify two types of networks: homophily and heterophily. Here, we use edge homophily \cite{zhu2020beyond} as a metric to measure the proportion of connections between nodes with the same label in the graph, defined as follows:
\begin{equation}
        \mathcal{H} = \frac{\left| \{ e_{ij} \mid e_{ij} \in \mathcal{E} \wedge {\bf Y}_i = {\bf Y}_j \} \right|}{\left| \mathcal{E} \right|},
\end{equation}

The homophily rate ranges from 0 to 1. A large $\mathcal{H}$ value $(\mathcal{H} \to 1)$ indicates the graph is homophilous, i.e., the target node and its neighboring nodes are likely to belong to the same class. On the contrary, a small $\mathcal{H}$ value  $(\mathcal{H} \to 0)$ indicates the graph is low in homophily, or what we refer to as heterophily.

In the experiment, our goal is to validate the effectiveness of the proposed GRAIN method on different types of datasets, particularly heterophilous datasets. Table \ref{compare2} displays the classification accuracy results on eight datasets, including three homophily and five heterophily, where the bold results represent the best scores in each dataset column. The last column of the table indicates the improvement of the proposed method compared to each row's method. In addition, we present the results of comparative experiments on five more datasets in the Appendix C.2. From the table, we can draw the following conclusions:

\begin{figure}[t]
    \centering
    \includegraphics[width=8.1cm]{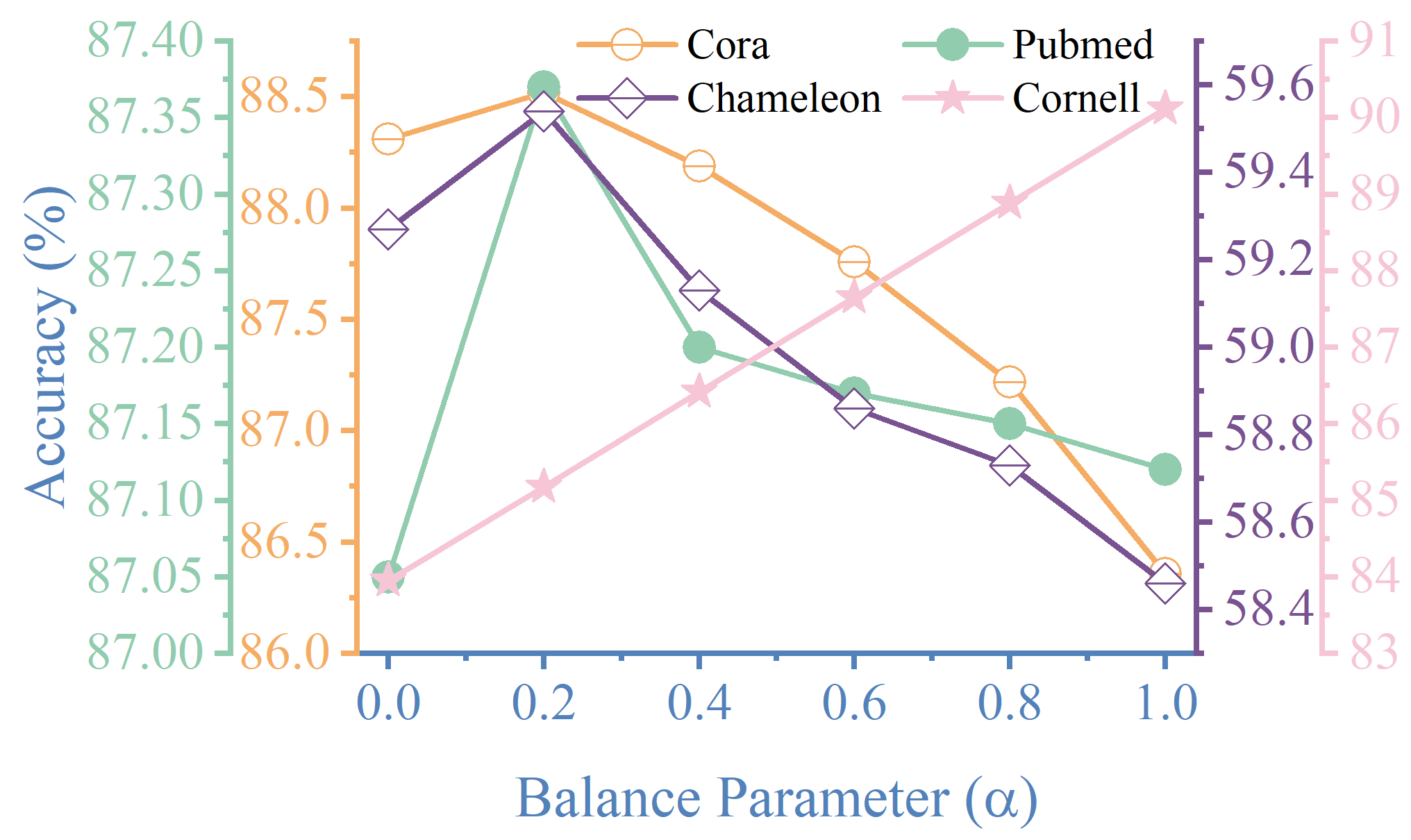}
    \caption{Analysis of balancing parameters ($\alpha$) for different datasets, where $\alpha$ influences the proportion of aggregation chosen for coarse- and fine-grained information.}
    \label{line}
\end{figure}

\begin{figure*}[t]
    \centering
    \includegraphics[width=14.9cm]{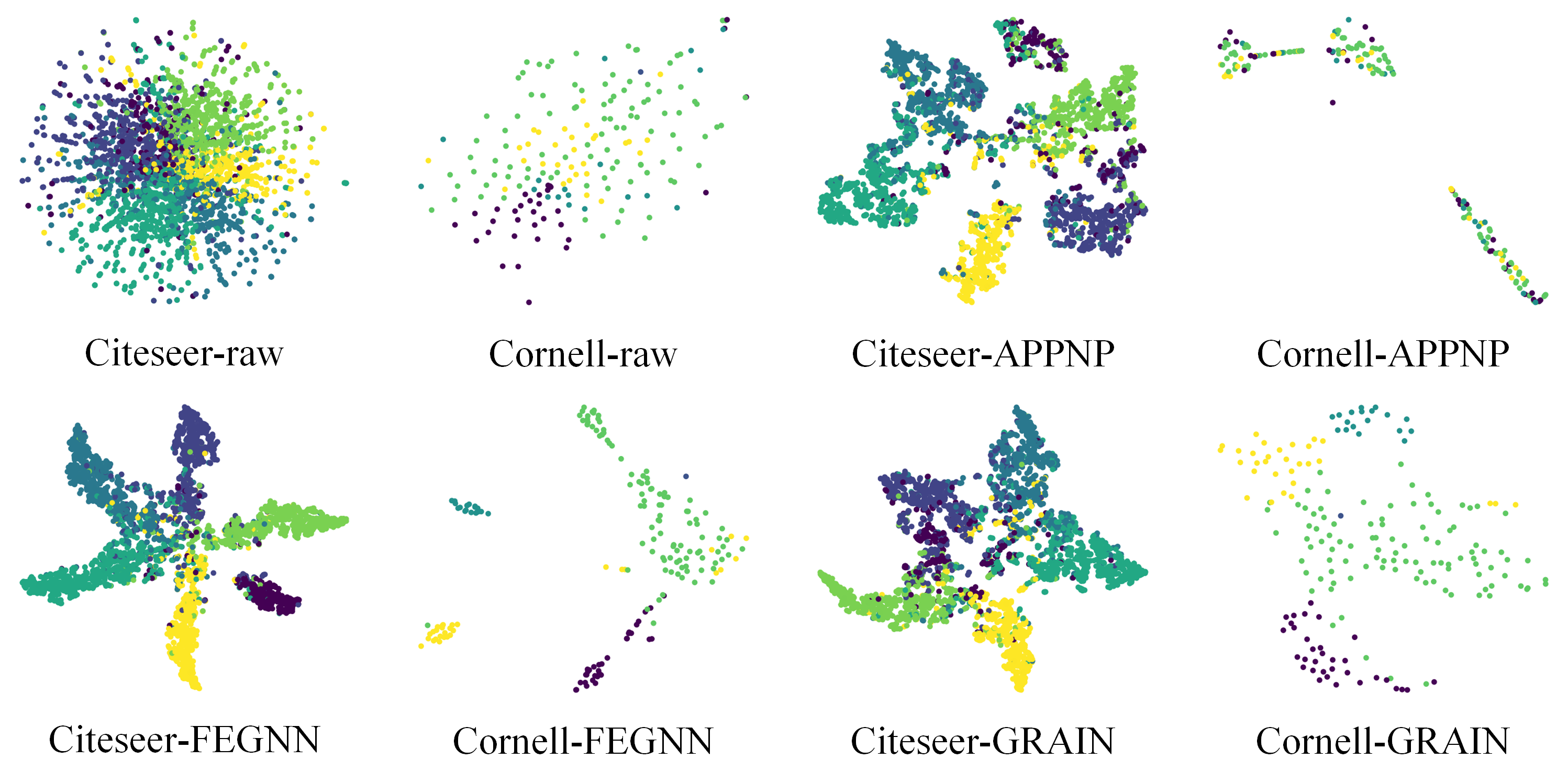}
    \caption{ The visualization of classification results of the proposed model for different datasets.} 
    \label{visualize}
\end{figure*}

1) MLP, which relies solely on node features, performs unexpectedly well on highly heterophilous data, surpassing three typical GNNs and homophily-generalized models. This outcome is likely due to MLP’s exclusion of structural information, thereby avoiding the noise introduced by aggregating heterophilous nodes. This suggests that simply employing heterophilous information without careful design  does not necessarily result in high-quality representations.
Moreover, the MLP's disregard for structural information leads to the poorest performance on homophilous datasets, underscoring the necessity of incorporating structural features via effective node representation learning.

2) Models designed for homophily also perform well on small-scale heterophilous datasets, mainly due to their enhanced node representations, which take into account more comprehensive information such as multi-hop neighbor information and tailoring features. However, their performance decreases on large-scale heterophilous datasets. These GNN methods heavily rely on the homophily assumption, and the aggregation process introduces noise due to the interference of heterophilous features, resulting in poor classification performance.

3) The methods for heterophilous networks, such as GNN-SATA, have shown relatively good results, especially on the Actor dataset. However, these methods may not consistently outperform others in homophilous networks. This could be because these methods are tailored for handling intricate relationships in heterophilous networks, such as conflicting node labels and dissimilar feature distributions. 

4) Notably, GRAIN demonstrates outstanding performance across heterophilous and homophilous datasets. Compared to the best baseline, it achieves an overall performance improvement of 2.36\%, and perform best across almost all datasets. This is primarily because our method considers different granularities of information and implicit features during the aggregation process, focusing on enhancing node representations. By considering different granularities of information, the model effectively learns structural information and avoid interference of heterophilous features, resulting in better embedding representations. In the Appendix C.3, we further ablate the effects of different information on the model with more types of RL methods. 


\subsection{Sensitivity Analysis}
\label{sensitive}

This section analyzes the hyperparameter $\alpha$ in the constructed aggregation function. The role of parameter $\alpha$ is to assess the importance of granularity information during the aggregation process. A smaller $\alpha$ indicates that coarse and fine-grained information has a greater impact on the representation, aiding in smoother node embeddings. The parameter $\alpha$ range is between 0 and 1, and we conduct a comparative analysis using equal intervals.

As illustrated in the line graph of Figure \ref{line}, it is evident that variations in $\alpha$ significantly impact the model across both homophily (Cora and Pubmed) and heterophily (Chameleon and Cornell).
Whether in heterophilous or homophilous settings, the model performs relatively better when $\alpha$ is set to 0.2. This suggests that granularity information plays a crucial role in achieving smooth node representations, primarily because it allows for the aggregation of more homophilous data within a certain range. However, when $\alpha$ is set too low, the critical features of the node itself may not be effectively utilized, leading to a decline in performance. Additionally, incorporating excessive coarse-grained information in the smaller Cornell network may introduce aggregation noise from heterogeneous data.


This phenomenon arises primarily due to the disparity in correlation between topology and node attributes, where even nodes with similar features may not be connected. The proposed aggregation function effectively integrates information across different granularities, preventing under-smoothing in node representations and ensuring more consistent and accurate embeddings in the latent space.


\subsection {Visualization of Learned Representations}

To intuitively validate the performance of the proposed model in node classification tasks, we employed the t-SNE method to visualize the node embeddings in a two-dimensional space, as depicted in Figure \ref{visualize}. In these visualizations, nodes with the same color belong to the same category. The results clearly illustrate that our proposed GRAIN model significantly outperforms the raw data and other methods in learning more distinct and well-separated embeddings for different categories. Specifically, GRAIN excels in capturing the inherent structure of the graph, as evidenced by the formation of clear, consistent clusters that are well-aligned with the underlying categories. This demonstrates GRAIN’s superior capability in learning meaningful and expressive graph representations, which translates into enhanced performance in node classification tasks. The sharp separation and distinct clustering of nodes in the two-dimensional space visually affirm that GRAIN is highly effective in distinguishing between categories.

\section{Related Works}
Several studies \cite{wang2020gcn,luan2021heterophily,yang2024graph} have focused on addressing heterophily in GNNs.
For instance, \citeauthor{abu2019mixhop} \shortcite{abu2019mixhop} recognizes that popular GNNs fail to learn general neighborhood mixing relationships, MixHop addresses this issue by mixing features of neighbors at different distances.
Geom-GCN \cite{pei2020geom} maps the graph to a continuous latent space via node embeddings, then defines geometric relationships and constructs structural neighbors for aggregation to handle heterophily.
H$_{2}$GCN \cite{zhu2020beyond} adapts to heterophilous networks by designing ego- and neighbor-embedding separation, higher-order neighbors, and combination of intermediate representations.
\citeauthor{yang2022graph} \shortcite{yang2022graph} iteratively updates representations of topology and attributes by simultaneously capturing semantic information and removing common information, thereby improving performance on heterophilous data.
\citeauthor{luan2022revisiting} \shortcite{luan2022revisiting} proposes an Adaptive Channel Mixing (ACM) framework, which adaptively utilizes aggregation, diversification, and identity channels to intelligently extract richer localized information and adapt to the heterogeneity of different nodes.
Ordered-GNN \cite{song2023ordered} proposes an effective message-passing strategy, using specific blocks of neurons for messages passed within specific hops to address heterophily.

These methods offer effective techniques for tackling heterophily in graph neural networks, advancing research, and providing new insights for managing heterogeneous relationships in complex networks. However, they often overlook the impact of aggregation granularity on node representation and fail to consider the implicit information between nodes in the graph. In contrast, our proposed GRAIN achieves multi-granularity information aggregation, resulting in smoother representations and providing a more generalized solution for heterophilous graphs.

\section{Conclusion}

In this paper, we introduce the GRAIN model to address the challenges posed by heterophilous graphs, which can achieve smoother node embedding by aggregating multi-view information.
Our approach effectively explores different granularity levels information, ensuring that both fine- and coarse-grained details are incorporated into the node embeddings, resulting in a richer representation.
Further, by incorporating the implicit relationships between distant nodes, GRAIN enhances the graph's contextual understanding, resulting in smoother and more accurate node embeddings.
We demonstrate the effectiveness of GRAIN in extensive comparative experiments on 13 datasets.

While our approach significantly improves training efficiency, substantial computational challenges remain when scaling to much larger networks. 
The iterative training required to learn the appropriate information granularity for different nodes can be resource-intensive.
Additionally, the next state is randomly selected in the process, which does not account for optimality guarantees.
In future work, we aim to develop more scalable optimization techniques to better manage complex graphs. Additionally, we plan to integrate more nuanced information, such as multi-scale and temporal features, via advanced aggregation functions. 

\section{Acknowledgments}

This work is partially supported in part by the National Natural Science Foundation of China (No. 62476110, No. U2341229); the National Key R\&D Program of China (No. 2021ZD0112500); the Key R\&D Project of Jilin Province (No. 20240304200SF); and the International Cooperation Project of Jilin Province (No. 20220402009GH).

\appendix
\section{A \quad Implementation details}

To develop GRAIN, we model graph representation learning as a Markov Decision Process (MDP) and use deep reinforcement learning to solve MDPs.

\subsection{A.1 \quad Markov Decision Process}

The MDP is a mathematical framework used for modeling decision-making problems, particularly suitable for situations involving uncertainty and dynamic environments. MDP describes the evolution of a system through states ($s$), actions ($a$), transition probabilities ($p$), and rewards ($r$). Here, $s$ represents the state of the system at a specific point in time, depicting the current situation or environment, with the finite set of states denoted as $\mathcal{S}$. The action $a$ represents the operations or decisions that the decision-maker can choose in each state, with the finite set of actions denoted as $\mathcal{A}$. The transition probability $p$ indicates the likelihood of the system moving to the next state, given the current state and action. The transition probability function $\mathcal{P}(s^\prime \mid s, a)$ describes the probability of transitioning from state $s$ to state $s\prime$ after taking action $a$. The reward $r$ represents the return obtained after executing an action. The reward function $\mathcal{R}(s, a)$ denotes the immediate reward received after performing action $a$ in state $s$.
Through these elements, MDP provides a systematic method for decision-makers to select the optimal policy ($\pi$) that maximizes cumulative rewards. The policy $\pi(a \mid s)$ describes the probability of choosing action $a$ in state $s$. The objective of an MDP is to find the optimal policy $\pi^\ast$ that maximizes long-term cumulative rewards. 

By modeling the graph representation learning problem as an MDP, we can systematically analyze and optimize this process, achieving better decision-making results in dynamic and uncertain graph network environments. The following sections will discuss how to express the graph representation learning process as a MDP. Specifically, we elucidate the essential components of the MDP in the context of graph representation learning, including states, actions, and rewards.

\subsection{A.2 \quad Twin Delayed Deep Deterministic Policy Gradient}

Twin Delayed Deep Deterministic Policy Gradient (TD3) \cite{fujimoto2018addressing} employs two critic networks with identical architectures. To calculate the target value, the algorithm selects the minimum value from the two critic networks to assess the next state-action value, which helps reduce the overestimation problem. The process is as follows:
\begin{equation}
        y = r + \gamma \cdot {min}_{i=1,2} Q_{\theta^\prime_i} \left( s^\prime, a^\prime \right),
        \label{target}
\end{equation}
where $\gamma$ denotes the discount factor, $\theta^\prime$ denotes the parameters of target the critic network, $Q$ refers to the Q-value estimated by the target critic network, $s^\prime$ and $a^\prime$ are the next state and action, respectively. The action $a^\prime$ under the state $s^\prime$ is computed using the target actor network $\left( \pi^\prime \right)$ as $a^\prime = \pi^\prime \left( s^\prime \mid \phi^\prime \right)$, where $\phi^\prime$ represents the parameters of the target actor network. However, deterministic policy learning is highly susceptible to function approximation errors, so noise $\epsilon$, a bounded random number following a normal distribution, is added to the target action $a^\prime$.
Finally, a gradient descent algorithm minimizes the error $\mathcal{L}$ between the estimated and target values, thereby updating the parameters in the critic1 and critic2 networks.
\begin{equation}
    \mathcal{L} = \left( Q_{\theta_i} \left(s, \pi_\phi \left( s \right) \right) - y \right)^2   (i=1,2)
    \label{loss}
\end{equation}

TD3 employs delayed updates to prevent the actor network from oscillating. Therefore, the update frequency of the policy network should be lower than that of the critic networks. The gradient expressions for the actor network parameters are computed using the following equations:
\begin{equation}
    \nabla_\phi J\left(\phi\right) = \mathbb{E}_{s \backsim \mathcal{D}} \left[ \nabla_a \ Q_{\theta_1} \left( s,a \right) \mid _{a=\pi_\phi\left(s\right)} \nabla_\phi \pi_\phi(s) \right]
\end{equation}
where $\mathcal{D}$ represents the replay buffer, $\phi$ is the parameters of the actor network, $\pi_\phi(s)$ denotes the action determined by the actor network. TD3 ensures that the actor network is updated conservatively by using this update rule, leading to more stable and reliable learning.

In addition to reducing the update frequency, soft updates should be used to update the target and policy networks. This is done using the following equations:
\[ \theta_i^\prime \gets \tau \theta_i + (1-\tau)\theta_i^\prime  \]
\[ \phi^\prime \gets \tau\phi + (1-\tau)\phi^\prime,  \]
where $\tau \in \left( 0,1 \right)$ is a learning rate that controls the extent to which the old target network parameters and the new corresponding network parameters are averaged. By applying soft updates, the target networks are updated smoothly, contributing to the stability of the training process. The process of execution of the algorithm is presented in Algorithm \ref{TD3}.

\begin{algorithm}
    \caption{TD3}
    \label{TD3}
    Initialize the critic networks $Q_{\theta_i} (i=1,2)$ and the actor network $\pi_\phi$ with random parameters \\
    Initialize the target network parameters $\theta^\prime_i \gets \theta_i (i=1,2)$ and $\phi^\prime \gets \phi$ \
    \begin{algorithmic}[1] 
        \FOR{$t = 1$ {\bf to} $T$}
        \STATE Calculate actions and add noise $a_t \gets \pi_\phi(s_t) + \epsilon$, where $\epsilon \sim N (0, \sigma)$
        \STATE Obtain reward $r_t$ and next state $s_{t+1}$
        \STATE Store $(s_t, a_t, r_t, s_{t+1})$ into \emph{Buffer}  \\
        \STATE Sample $batches$ of transitions $(s_t, a_t, r_t, s_{t+1})$ from \emph{Buffer}
        \STATE $\hat{a} \gets \pi_{\phi^\prime}(s_{t+1} + \epsilon)$, $\epsilon \sim clip\left(N (0, \sigma), -c, c \right)$
        \STATE $y \gets r + \gamma \cdot {min}_{i=1,2} Q_{\theta^\prime_i} \left( s_{t+1}, \hat{a} \right)$
        \STATE Calculate the $Q$ loss function: \\
        $\mathcal{L} = \left( Q_{\theta_i} \left(s, \pi_\phi \left( s \right) \right) - y \right)^2   (i=1,2)$ 
        \STATE Update critic network parameters $\theta_i (i = 1, 2)$
        \IF{$t$ mod $delay$}
        \STATE Update $\phi$ by the deterministic policy gradient: \\
        $\nabla_\phi J\left(\phi\right) = \mathbb{E}_{s \backsim \mathcal{D}} \left[ \nabla_a \ Q_{\theta_1} \left( s,a \right) \mid _{a=\pi_\phi\left(s\right)} \nabla_\phi \pi_\phi(s) \right]$
        \STATE Update critic network parameters: \\
        $\theta_i^\prime \gets \tau \theta_i + (1-\tau)\theta_i^\prime$  \\
        $\phi^\prime \gets \tau\phi + (1-\tau)\phi^\prime$
        \ENDIF
        \ENDFOR
    \end{algorithmic}
\end{algorithm}

\section{B \quad Model Architecture}

The pseudocode for GRAIN training is shown in Algorithm {\ref{alg:algorithm}}, which takes the following steps. 1) we model graph representation learning as a Markov Decision Process (MDP), making the problem description and solution more explicit, i.e., lines 1 {-} 7 (Algorithm {\ref{alg:algorithm}}). 2) As shown in lines 9  {-} 18 (Algorithm {\ref{alg:algorithm}}), we employ deep RL algorithms to learn optimal strategies for solving the MDP. RL can explore different granularity levels of node information and consider implicit information between nodes, effectively integrating local and global information to enhance node embedding performance. 3) we design a novel node information aggregator that effectively combines multi-granularity and implicit information, improving the quality of node representations, which is shown in lines 20 {-} 24. The source code of GRAIN is available at \url{https://github.com/Songwei-Zhao/GRAIN}.

\subsection{B.1 \quad Experiment Settings and Hyperparameters}

We run all experiments in PyTorch, using a machine with 10-core Intel(R) Xeon(R) Gold 5218R CPU@2.10GHz, 64GB RAM, and an NVIDIA RTX 3090-24GB GPU. For the experiments, we make use of the same training/validation/test divisions as in the paper. GNN models minimize the cross-entropy in node training and employ the Relu activation function, the dropout rate is set to 0.5. \emph{Adam} optimizer is used to optimize all models. 
\begin{algorithm}[!h]
    \caption{ Granular and Implicit Graph Network}
    \label{alg:algorithm}
    \textbf{Input}: An undirected graph with node features and lables: $\mathcal{G}=(\mathcal{V}, \mathcal{E}, \bf {X}, \bf {Y})$; Training epoch: $T$; TD3 training step: $K$  \\
    Initialize all network parameters and $Buffer$ \\
    \textbf{Output}: $\bf \widehat{Y}$: Prediction Label
    \begin{algorithmic}[1] 
        \STATE Initialize parameters $\epsilon, \vartheta, \varphi$  \\
        // \emph{Training process of reinforcement learning based on graph representation learning}
        \STATE Sample a node with its features as state $s_0$
        \FOR{$t = 0$ {\bf to} $T$}
        \STATE Calculate actions and add noise $a_t \gets \pi_\phi(s_t) + \epsilon$, where $\epsilon \sim N (0, \mu)$
        \STATE $r_t \gets \frac{\varphi \cdot \sum_{l=t-\vartheta}^{t} \left[ \mathcal{F} \left(s_t,a_t\right) -\mathcal{F} \left(s_l,a_l\right)\right]} {\vartheta+1}$ 
        \STATE Sample next state $s_{t+1}$ from $a_t$-hop neighborhood
        \STATE Store $(s_t, a_t, r_t, s_{t+1})$ into \emph{Buffer}  \\
        \FOR{$step = 0$ {\bf to} $K$}
        \STATE Sample $batches$ of transitions $(s_t, a_t, r_t, s_{t+1})$ from \emph{Buffer}
        \STATE $\hat{a} \gets \pi_{\phi^\prime}(s_{t+1} + \epsilon)$, $\epsilon \sim clip\left(N (0, \sigma), -c, c \right)$
        \STATE $y \gets r + \gamma \cdot {min}_{i=1,2} Q_{\theta^\prime_i} \left( s_{t+1}, \hat{a} \right)$
        \STATE Calculate the $Q$ loss function: \\
        $\mathcal{L} = \left( Q_{\theta_i} \left(s, \pi_\phi \left( s \right) \right) - y \right)^2   (i=1,2)$ 
        \STATE Update critic network parameters $\theta_i (i = 1, 2)$
        \IF{$t$ mod $delay$}
        \STATE Update $\phi$ by the deterministic policy gradient: \\
        $\nabla_\phi J\left(\phi\right) = \mathbb{E}_{s \backsim \mathcal{D}} \left[ \nabla_a \ Q_{\theta_1} \left( s,a \right) \mid _{a=\pi_\phi\left(s\right)} \nabla_\phi \pi_\phi(s) \right]$
        \STATE Update critic network parameters: \\
        $\theta_i^\prime \gets \tau \theta_i + (1-\tau)\theta_i^\prime$  \\
        $\phi^\prime \gets \tau\phi + (1-\tau)\phi^\prime$
        \ENDIF
        \ENDFOR
        \ENDFOR   \\
        // \emph{Training GNNs with learned optimal policy}
        \STATE Obtain the learned optimal policy
        \FOR{$t = 0$ {\bf to} $T$}
        \STATE Obtain $a_t$ through the optimal policy
        \STATE ${\bf H}_i^{(l+1)}=\frac{1} {\langle a_t \rangle} \sum\nolimits_{k =1}^{\langle a_t \rangle} \left[ (1 - \alpha) \hat{{\bf A}}_i^k {\bf H}^{(l)} + \alpha {\bf H}_i^{(l)}  \right]  + $  \\
        $(a_t - \left\lfloor a_t \right\rfloor) \hat{{\bf A}}_i^{\langle a_t \rangle} {\bf H}_i^{(l)} + ( \left\lceil a_t \right\rceil - a_t ) \hat{{\bf A}}_i^{\langle a_t \rangle + 1} {\bf H}_i^{(l)}$   \\
        // \emph{Different granularity and implicit information aggregation}
        \ENDFOR 
    \end{algorithmic}
\end{algorithm}
For the comparison algorithm in the paper, we performed a hyperparameter search on the validation set. All important notations in this paper are summarized in Table \ref{re1}. 

\section{C \quad Additional Experiments}

\subsection {C.1 \quad Dataset Description}
\begin{table}
    \centering
    \begin{tabular}{ll}
        \toprule
        \toprule
        Symbols         &  Definitions     \\ 
        \toprule
        $\mathcal{G}$    & An undirected graph with node set and edge set  \\
        $\mathcal{V}; \mathcal{E}$      &  The set of node; The set of edge  \\
        $\bf A$     & The adjacency matrix of graph $\mathcal{G}$   \\
        $\bf D$     & The degree matrix of graph $\mathcal{G}$   \\
        $\bf X$           & The feature matrix of nodes  \\
        $\bf Y$   & The label matrix of nodes \\
        $\mathcal{N}_k(v)$   &  The $k$-th order neighbors of node $v$ \\
        $T$            & Model total training epochs  \\
        $\bf W$     & learnable weight matrix        \\
        $\sigma$            & Non-linear activation function  ReLU  \\
        \toprule
        $\epsilon$      & Random noise that obeys the normal distribution   \\
        $\vartheta$      & The historical step of reward      \\
        $K$       & TD3 total training step         \\
        $\mathcal{S}$     & The set of states   \\
        $\mathcal{A}$      & The set of actions   \\
        $s_t$     & The state in step $t$    \\
        $a_t$   & The action in the step $t$  \\
        $r_t$   & The reward in the step $t$  \\
        $Q_{\theta_i}$  & Critic networks to evaluate the state $s$ and action $a$    \\  
        $\pi_\phi$        & Actor network to generate $a$ for a state $s$   \\
        \toprule
        \toprule
    \end{tabular}
    \caption{Major notations and definitions. The top rows are for graph representation learning; the bottom rows cover reinforcement learning.}
    \label{re1}
\end{table}

\begin{table*}
    \centering
    \scalebox{1}{
    \begin{tabular}{c|c|ccccc}
        \hline
        \toprule
        Type      & Dataset        & \#Nodes  & \#Edges & \#Features  & \#Classes & $\mathcal{H}$    \\ 
        \toprule
        \multirow{6}{*}{Homophily}  & Pubmed   & 19,717  & 44,327       & 500    &3  & 0.80                 \\
        & Citeseer      & 3,327 & 4,732      & 3,703 & 6   & 0.74                 \\
        & Cora     & 2,708 & 5,278      & 1,433 & 7   & 0.81               \\
        & Computers   & 13,752 & 245,861      & 745    & 10  & 0.79                  \\
        & Photo    & 7,650     & 119,081          & 745      & 8   & 0.82                    \\
        & Coauthor CS    & 18,333     & 81,894          & 6,805      & 15   & 0.83                    \\    \toprule
        \multirow{7}{*}{Heterophily} & Film    & 7,600     & 33,544          & 931      & 5   & 0.37                    \\
        & Cornell  & 183     & 280          & 1,703      & 5   & 0.30                    \\
        & Actor  & 7,600     & 26,752          & 931      & 5   & 0.22                    \\
        & Wisconsin  & 251     & 466          & 1,703      & 5   & 0.21                    \\
        & Squirrel  & 5,201     & 198,493         & 2,089      & 5   & 0.22                    \\
        & Texas  & 183     & 295          & 1,703      & 5   & 0.11                    \\
        & Chameleon  & 2,277     & 31,421          & 2,325      & 5   & 0.23                    \\
        \hline
        \toprule
    \end{tabular}
}
    \caption{Datasets statistics, where $\mathcal{H}$ is defined as the fraction of edges connecting nodes with the same label.}
    \label{dataset}
\end{table*}

In the citation networks, nodes and edges represent papers and citations, respectively, with node labels indicating paper topics and features representing the bag-of-words of the papers. Computers and Photo are segments of the Amazon co-purchase graph, where nodes and edges correspond to products and frequent co-purchases. Node features are bag-of-words encoded product reviews, and labels denote product categories. Coauthor CS is derived from the Microsoft Academic Graph used in the 2016 KDD Cup challenge\footnote{
https://kddcup2016.azurewebsites.net/.}, where nodes represent authors and edges represent co-authorship of papers.

For heterophily data, we use seven network datasets \cite{pei2020geom}, including WebKB\footnote{
http://www.cs.cmu.edu/afs/cs.cmu.edu/project/theo-11/www/wwkb/.} (Cornell, Wisconsin, and Texas), Film, Actor \cite{tang2009social}, Squirrel, and Chameleon. In these network datasets, nodes and edges represent web pages and links, respectively, with node features represented by the bag-of-words of the web pages. The statistical properties of the datasets are outlined in Table \ref{dataset}.

\subsection {C.2 \quad Additional Comparative Experiment}

Table \ref{compare1} presents the performance experimental results of all methods on an additional five real-world datasets. We use classification accuracy as the comparison metric and highlight the best score in each dataset column in bold. To provide an intuitive understanding of the model performance improvements, we list the improvement degree relative to the comparison methods. From the table, we can observe that:

1) MLP, which only considers node features and ignores structural features, performs the worst in homophilous datasets but relatively strongly in heterophily. This suggests that leveraging structural features is crucial for improving the quality of node representations in homophilous data. However, in highly heterophilous environments, this approach may introduce noise into the embeddings, thereby negatively impacting classification performance.

2) Compared to typical GNN models (GCN, GAT, and GraphSAGE), homophily-generalized models (MixHop, APPNP, and GCNII) achieve better experimental results. For instance, in the Coauthor CS dataset, MixHop, APPNP, and GCNII achieve classification accuracies of 95.24\%, 94.79\%, and 94.14\%, respectively, which are significantly better than the three typical GNN models. This is because they consider more comprehensive node representation information, such as multi-hop neighbor information and tailoring features. However, their performance in heterophily is generally poor because they rely on the homophily assumption for information aggregation, which overlooks the heterophilous scenario where connected nodes may have different attributes or labels.

3) Models designed for heterophilous networks often underperform in homophily because they overlook coarse and fine-grained information during the aggregation process. Additionally, while GNN-SATA achieves good results, it suffers from memory overflow issues, preventing it from being tested on large-scale datasets and limiting its generalization performance.

4) The proposed method, GRAIN, achieves the highest scores across all datasets, demonstrating its consistent ability to deliver excellent results across varying levels of homophily. This success is due to its capability to explore information at different granularities and capture implicit relationships between distant nodes. Furthermore, the proposed aggregator effectively integrates these factors during the aggregation process, enhancing the model's representation learning capability.

\subsection {C.3 \quad Comparison Experiment}

\begin{figure*}[t]
\centering 
    \includegraphics[width=14cm]{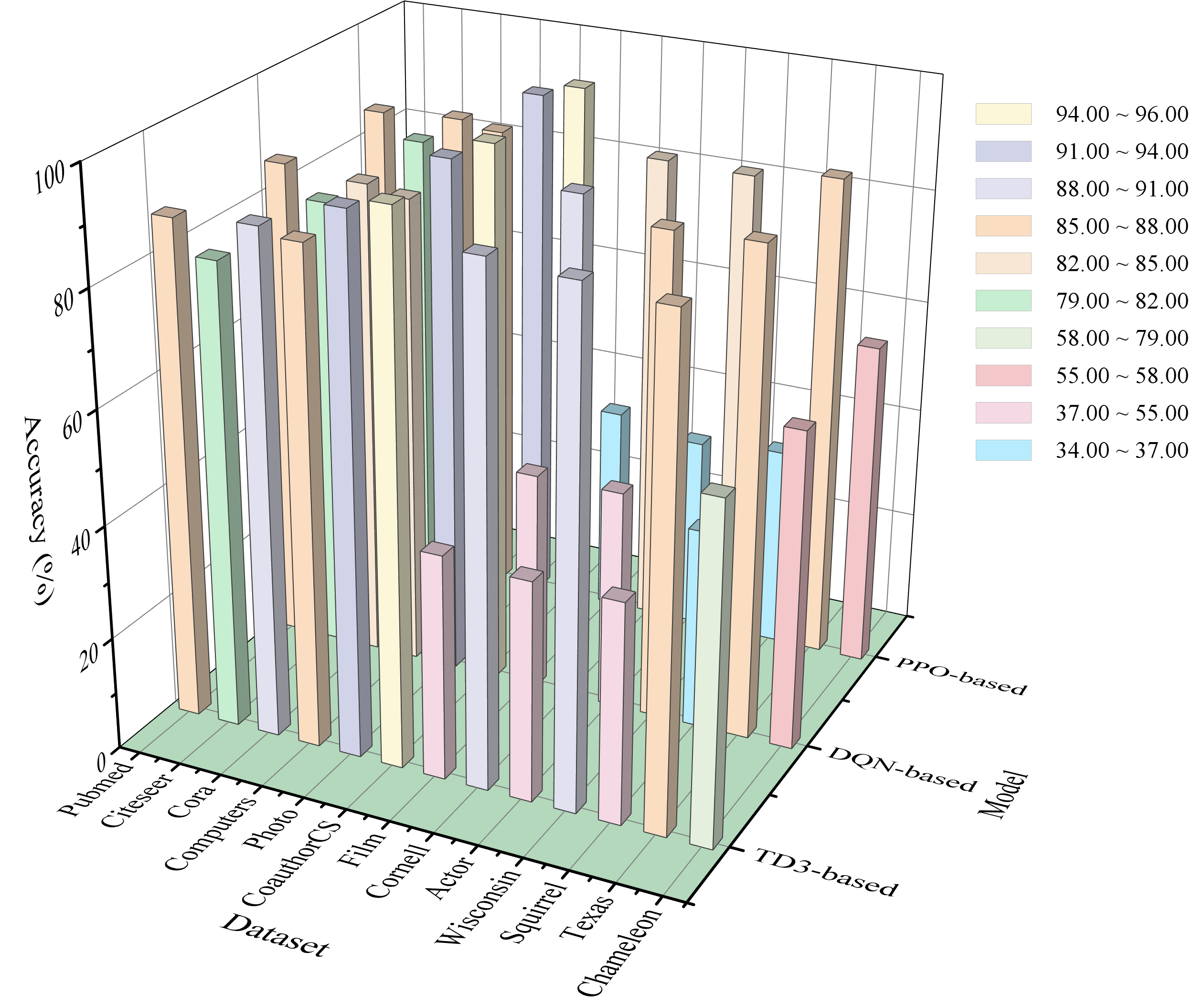}
    \caption{ Performance of the proposed method based on different reinforcement learning on a graph representation learning task.}
    \label{com}
\end{figure*}

Reinforcement Learning (RL) is a machine learning method in which an agent interacts with an environment and learns from feedback (rewards) to optimize decision-making strategies, thereby achieving goals in dynamic and uncertain environments. Currently, RL can be broadly categorized into three types: policy gradient methods, value-based methods, and actor-critic methods.

{\bf Policy Gradient Methods} do not rely on a value function but directly optimize the policy to maximize the probability of selecting the best action in a given state. These methods use parameterized policies and maximize cumulative rewards through gradient ascent. Common policy gradient methods include Proximal Policy Optimization (PPO) \cite{schulman2017proximal} and Deep Deterministic Policy Gradient (DDPG) \cite{lillicrap2015continuous}. Policy gradient methods can handle continuous action spaces and are suitable for high-dimensional and complex decision problems. However, they suffer from low sample efficiency, susceptibility to local optima, and slow convergence.
{\bf Value-based methods} estimate the value of each state or state-action pair to learn the optimal policy indirectly. The core idea of these methods is to explore a value function that evaluates the expected return of each state (or state-action pair). Typical value-based methods include Q-learning \cite{watkins1992q} and Deep Q-Network (DQN) \cite{mnih2013playing}. Value-based methods have high sample efficiency and fast convergence. However, they are challenging to extend to high-dimensional or continuous action spaces.
{\bf Actor-critic methods} combine the advantages of policy gradient and value-based methods. They consider the interaction between the approximations of the policy and the value updates. The actor is responsible for selecting actions, while the critic evaluates the current policy's value. The critic guides the actor's policy updates through the value function (state-value function or advantage function). Common actor-critic methods include Twin Delayed Deep Deterministic Policy Gradient (TD3) \cite{fujimoto2018addressing} and Asynchronous Advantage Actor-Critic (A3C) \cite{mnih2016asynchronous}. Actor-critic methods inherit the strengths of both approaches and can handle high-dimensional and continuous action spaces effectively.
To meet the technical requirements proposed in this study, we aim to enhance the quality of embeddings in graph representation learning by incorporating multi-granularity information and implicit relationships between distant nodes during the aggregation process. Based on the task requirements, we intend to utilize reinforcement learning to select granularity information for the target nodes. Since implicit information needs to be considered, the action space must be extended to the continuous domain, making actor-critic methods suitable for our framework.

\begin{table*}[!htbp]
    \centering
    \begin{tabular}{c|c|ccccc|c}
        \hline
        \toprule
        Type           &Method      & Film                & Squirrel                   & Computers               & Photo             & Coauthor CS    & Improve ($\downarrow$)           \\
        \toprule
        \multirow{4}{*}{Tradition}   &MLP   & 34.92          & 27.86                & 80.44          & 86.07          & 92.85   & 10.28      \\ 
        &GCN   & 32.99          & 28.45               & 82.03          & 92.84          & 93.60     & 7.69          \\
        &GAT   & 33.12          & 30.36               & 85.67          & 93.06          & 93.61      & 5.79          \\
        &GraphSAGE   & 34.15          & 29.28               & 84.23        & 93.41         & 94.03     & 6.02       \\  \toprule
        \multirow{3}{*}{Homophily}   &MixHop & 33.77          & 31.63          & 86.11          & 93.24          & 95.24      & 4.49    \\  
        &APPNP & 34.45         & 28.24            & 85.35          & 93.54          &94.79        & 5.62  \\
        &GCNII & 35.02         & 27.47               & 86.41          & 93.59          & 94.14      & 5.54    \\   \toprule
        \multirow{5}{*}{Heterophily}   &GPR-GNN & 36.53         & 33.59              & 86.31          & 93.64          & 94.46       & 3.12    \\ 
        &GGCN  & 37.81      &38.75             & 86.98        &90.26           & 95.39     & 1.74 \\
        &ACM-GNN & 39.33         & 38.55            & 86.56          & 93.66          & 95.37      & 0.5      \\
        &FE-GNN & 39.21          & 35.73             & 87.05          & 93.31          & 95.33      & 1.32      \\
        &GNN-SATA & 39.48        & 38.69            & OOM          & 93.56          & OOM         & 0.46    \\   \toprule
        Ours        &GRAIN   & \textbf{39.77}           & \textbf{38.96}               & \textbf{87.16}          & \textbf{93.79}          & \textbf{95.59}    & --        \\ 
        \hline
        \toprule
    \end{tabular}
    \caption{Experimental results: the average test classification accuracy (\%) across all methods on six homophily datasets. Improve ($\downarrow$) denotes how much the GRAIN improves performance relative to that row's method. We highlight the optimal results on each dataset in bold. OOM refers to ``out-of-memory''.}
    \label{compare1}
\end{table*}

To validate our approach, we conducted comparative experiments using representative algorithms from the three methods above, with the results presented in Figure \ref{com}. The figure shows that the actor-critic method outperforms the other two methods across most datasets. However, the Actor dataset is slightly poorly compared to the value-based method. This indicates the effectiveness of our proposed approach.
First, the value-based method is designed for discrete action spaces. Thus, it only considers the impact of different granularity information on node representations and neglects the role of implicit information, making its performance inferior to the other two methods but still superior to traditional GCN. Second, the policy gradient method performs less than the actor-critic method due to its low sample efficiency and tendency to get trapped in local optima.

In conclusion, incorporating different granularity information and implicit relationships between nodes into the node aggregation process is irreplaceable in enhancing node embeddings. By integrating these elements, we can improve the smooth representation of nodes, significantly enhancing the performance of various tasks.


\end{document}